%% file: main.tex
\documentclass[10pt,twocolumn,letterpaper]{article}

 \usepackage{cvpr}              

\usepackage{booktabs}
\usepackage{multirow}
\usepackage{pifont}
\usepackage{xcolor}
\definecolor{darkgreen}{rgb}{0.0, 0.5, 0.0}

\definecolor{cvprblue}{rgb}{0.21,0.49,0.74}
\usepackage[pagebackref,breaklinks,colorlinks,allcolors=cvprblue]{hyperref}

\def\confName{CVPR}
\def\confYear{2026}

\title{Adaptive Geodesic Conformal Prediction for Egocentric Camera Pose Estimation

}

\author{Aishani Pathak\\
Arizona State University\\
{\tt\small ajpathak@asu.edu}
\and
Hasti Seifi\\
Arizona State University\\
{\tt\small hasti.seifi@asu.edu}
}

\begin{document}
\maketitle
\input{sec/0_abstract}    
\input{sec/1_intro}
\input{sec/2_formatting}

\input{sec/3_experiments}
{
    \small
    \bibliographystyle{ieeenat_fullname}
    \bibliography{main}
}

 \end{document}

%% file: sec/0_abstract.tex
\begin{abstract}

Egocentric pose estimation for Augmented Reality (AR) and assistive devices requires 
not just accurate predictions but guaranteed uncertainty 
regions. Conformal prediction (CP) provides such guarantees 
without retraining, but we show that standard CP with a single 
fixed threshold achieves nominal 90\% overall coverage while 
covering only $\sim$60\% of the hardest 25\% of frames (Q4) --- 
a $\sim$30 percentage-point conditional coverage gap consistent 
across 12 participants, 3 predictors, and 3 horizons (108 
evaluations) on EPIC-Fields. We further show that a geodesic 
SE(3) nonconformity score identifies physically harder frames 
than Euclidean scoring, with only 15--26\% Q4 overlap and 
2--3$\times$ higher ground-truth camera displacement for 
geodesic Q4 frames. To close the coverage gap, we propose 
DINOv2-Bridge adaptive CP: a two-stage difficulty estimator 
trained on a single source participant that transfers 
cross-participant without any images at test time, improving 
Q4 coverage from $\sim$0.75 to $\sim$0.93 while maintaining 
overall coverage at the 90\% target.
\end{abstract}

%% file: sec/1_intro.tex
\section{Introduction}
\label{sec:intro}

Wearable cameras in AR glasses and assistive devices must know 
where they are pointing at every moment. This camera localization, 
called egocentric pose estimation, is essential for AR systems to 
overlay guidance, track objects, and support real-time 
decision-making. For safety-critical applications such as surgical 
assistance, industrial task guidance, and eldercare support, a 
single best-guess location is not enough. The system must also 
know how reliable that guess is. While some visual odometry (VO) 
methods produce heuristic confidence scores or learned covariance 
estimates, these lack formal statistical coverage guarantees: they 
do not ensure that the true pose falls within the reported 
uncertainty region with a specified probability. Conformal 
prediction (CP)~\cite{vovk2005,angelopoulos2022} addresses this 
by wrapping any pre-trained model with a distribution-free, 
model-agnostic uncertainty region --- no retraining required --- 
guaranteed to contain the true pose with probability at least 
$1-\alpha$. CP first runs on a held-out calibration set to find 
a threshold $\hat{q}$ at the $(1-\alpha)$ quantile of nonconformity 
scores (scalar prediction errors); at test time, it reports all 
outputs whose error falls below $\hat{q}$.

CP has recently been applied to pose estimation across several 
domains. \citet{yang2023} apply CP to 6D object pose via keypoint 
prediction sets with Euclidean nonconformity scores, where keypoint 
uncertainty is translated into pose uncertainty by using the known 
geometric relationship between image keypoints and the object's 3D 
position. \citet{closure2024} extend this to geodesic uncertainty 
sets on SE(3) (the Lie group of rigid-body motions combining 
rotation and translation) for static object pose. \citet{stutts2023} 
present a conformalized VO framework for edge robotics using per degree-of-freedom Euclidean prediction intervals. 
Despite these advances, three critical gaps remain: (i)~none address 
egocentric camera localization from continuous video; (ii)~all 
evaluate only \textit{marginal} coverage --- the average guarantee 
over all test inputs, which can mask systematic failures on specific 
subgroups such as fast-moving frames --- without examining whether 
coverage holds for the hardest inputs specifically; and (iii)~none 
consider cross-participant generalization, where a system must 
provide reliable uncertainty for people it has never calibrated on.

The choice of nonconformity score is an important factor in CP. Camera poses lie 
on SE(3), where Euclidean distance distorts true prediction error: 
a small Euclidean difference in rotation matrices can correspond to 
a large physical rotation. \citet{shahbazi2026} propose using 
geodesic distance for geomagnetic forecasting on $\mathbb{S}^2$ 
(the 2-sphere, manifold of unit-norm vectors in $\mathbb{R}^3$), 
but do not address SE(3) or cross-participant generalization.

In this work, we apply split CP with a geodesic SE(3) nonconformity 
score to egocentric camera pose estimation on 
EPIC-Fields~\cite{epicfields2023} and demonstrate the need for 
adaptive CP. Even with the geometrically correct score, a single 
fixed threshold --- which we term \textit{standard CP} --- achieves 
nominal 90\% overall coverage but covers only $\sim$60\% of the 
hardest 25\% of frames (Q4). This $\sim$30 percentage-point 
conditional coverage gap is consistent across 12 participants, 3 
pose predictors, and 3 prediction horizons (108 evaluations), 
revealing that the failure is not a property of any particular 
predictor or score, but of using a single threshold across frames 
of wildly varying difficulty. We further show that geodesic and 
Euclidean Q4 frame sets overlap only 15--26\% for feature-matching 
predictors, and geodesic Q4 frames have 2--3$\times$ higher camera 
translation --- confirming geodesic scoring identifies physically 
harder frames. To close the gap, we propose adaptive CP for 
egocentric pose estimation. Standard CP assigns a single fixed 
threshold to every test input regardless of difficulty; adaptive CP 
instead assigns a per-frame difficulty-weighted threshold, expanding 
prediction regions for hard frames and shrinking them for easy ones. 
Implementing adaptive CP requires a per-frame difficulty score 
$\sigma_i$. We estimate $\sigma_i$ via DINOv2-Bridge: a model 
trained on a single source participant (P01, 56K frames) that 
predicts frame difficulty from visual appearance using DINOv2 
features, then distills this into pose-kinematic features for 
image-free inference at test time. We apply DINOv2-Bridge to the 
constant-velocity predictor because its residuals are driven by 
motion kinematics --- velocity and acceleration --- which align 
naturally with the kinematic features the Bridge uses to estimate 
difficulty. DINOv2-Bridge improves coverage on the hardest frames 
(Q4) from $\sim$0.75 to $\sim$0.93 across two cross-participant 
splits, each with four unseen test participants, while maintaining 
overall coverage at the 90\% target.

\noindent\textbf{Contributions:}
\begin{itemize}
  \item[\textbf{(i)}] First characterization of the conditional 
  coverage gap for egocentric SE(3) pose CP on EPIC-Fields --- 
  $\sim$30pp Q4 shortfall consistent across 12 participants, 
  3 predictors, 3 horizons (108 evaluations), with Q1--Q3 at 
  nominal coverage and Q4 at $\sim$60\%.
  \item[\textbf{(ii)}] Geometric analysis showing geodesic SE(3) 
  scoring correctly identifies physically hard frames while 
  Euclidean scoring does not, with only 15--26\% Q4 overlap.
  \item[\textbf{(iii)}] DINOv2-Bridge adaptive CP for the 
  constant-velocity predictor, trained on one participant 
  (P01, 56K frames), generalizing cross-participant without 
  retraining and closing Q4 from $\sim$0.75 to $\sim$0.93.
\end{itemize}

%% file: sec/2_formatting.tex
\section{Geodesic Conformal Prediction on SE(3)}
\label{sec:method}

\noindent\textbf{Split conformal prediction.}
Let $\{(x_i, y_i)\}_{i=1}^{n}$ be a held-out calibration set where 
$y_i \in \text{SE}(3)$ is a ground-truth camera pose and $\hat{y}_i$ 
is the predictor's estimate. Split CP computes a nonconformity score 
$s_i = d(\hat{y}_i, y_i)$ for each calibration point, where $d$ is 
a distance function on the output space. The threshold is set as:
\begin{equation}
    \hat{q} = \text{Quantile}\!\left(\{s_i\}_{i=1}^{n},\, 
    \tfrac{\lceil(1-\alpha)(n+1)\rceil}{n}\right)
\end{equation}
At test time, the prediction set $\mathcal{C}(x) = \{y : d(\hat{y}, y) 
\leq \hat{q}\}$ contains the true pose with probability at least 
$1-\alpha$ over the calibration randomness~\cite{angelopoulos2022}.

\noindent\textbf{Geodesic nonconformity score on SE(3).}
Standard CP applied to pose estimation typically uses Euclidean or 
Frobenius distance as $d$, implicitly treating SE(3) as flat space. 
We instead use the geodesic distance on SE(3), which correctly 
accounts for the Riemannian geometry of rotation:
\begin{equation}
    d_{\text{geo}}(\hat{y}, y) = \sqrt{W_R \,\|\log(R^\top \hat{R})\|^2 
    + W_T \,\|t - \hat{t}\|^2}
\end{equation}
where $R, \hat{R} \in \text{SO}(3)$ are true and predicted rotations, 
$t, \hat{t} \in \mathbb{R}^3$ are translations, $\log(\cdot)$ is the 
SO(3) logarithmic map, and $W_R, W_T$ are scale weights. This score 
correctly measures rotation error as arc length on the unit sphere 
rather than chord length in matrix space.

\noindent\textbf{Adaptive CP via DINOv2-Bridge.}
Standard CP assigns the same threshold $\hat{q}$ to every test 
input, regardless of difficulty. Following the normalized score 
approach of~\citet{romano2019}, we replace $s_i$ with difficulty-
weighted scores $\tilde{s}_i = s_i / \sigma_i$, where $\sigma_i > 0$ 
is a predicted difficulty for input $i$. The adaptive threshold 
$\tilde{q}$ is computed on $\{\tilde{s}_i\}$ and applied as 
$\mathcal{C}(x) = \{y : d_{\text{geo}}(\hat{y}, y) \leq \tilde{q} 
\cdot \sigma\}$, yielding larger regions for hard inputs and smaller 
ones for easy inputs. We estimate $\sigma$ via a two-stage Bridge: 
a DINOv2-MLP trained on a source participant (e.g., P01) maps visual features 
to difficulty scores, and a lightweight Bridge multilayer perceptron (MLP) learns to predict 
these difficulty scores from pose features alone, enabling difficulty 
estimation at test time without any image access.

%% file: sec/3_experiments.tex
\section{Experiments}
\label{sec:experiments}

\noindent\textbf{Setup.}
We evaluate on EPIC-Fields~\cite{epicfields2023}, which provides 
ground-truth camera trajectories using COLMAP ~\cite{pan2024globalstructurefrommotionrevisited}(Structure-from-Motion pipeline for multi-view 3D reconstruction) for EPIC-KITCHENS~\cite{damen2022} 
participants recorded in kitchen environments. 
This benchmark is well-suited for our study because COLMAP provides
millimeter-accurate ground-truth poses across diverse kitchen
activities, and the multi-participant structure enables both
within-participant and cross-participant evaluation of generalization.
We use 12 participants 
(P01--P12), 3 pose predictors namely, constant-velocity (const-vel), 
LightGlue~\cite{lindenberger2023}, and MonoDepth2~\cite{godard2019} 
and 3 prediction horizons $k \in \{10, 20, 30\}$ frames, 
with $\alpha = 0.10$ (targeting 90\% coverage) throughout.

\noindent\textbf{Evaluation protocol.}
We consider two settings. In the \textit{within-participant} setting 
(Sec.~\ref{sec:gap}--\ref{sec:geo_vs_euc}), each participant's 
sequences are split chronologically: the first 50\% form the 
calibration set and the last 50\% form the test set. In the 
\textit{cross-participant} setting (Sec.~\ref{sec:adaptive}), 
one participant's full sequence is used for calibration and 
a disjoint set of participants forms the test set.

\begin{table*}[t]
\centering
\caption{Within-participant Q4 coverage ($\alpha=0.10$). Target: 0.90. 
All 108 participant-predictor-prediction horizon combinations fall 
below target.}
\label{tab:story1}
\setlength{\tabcolsep}{4pt}
\small
\begin{tabular}{l ccc ccc ccc}
\toprule
& \multicolumn{3}{c}{$k=10$} & \multicolumn{3}{c}{$k=20$} & \multicolumn{3}{c}{$k=30$} \\
\cmidrule(lr){2-4}\cmidrule(lr){5-7}\cmidrule(lr){8-10}
\textbf{P} & \textbf{LG} & \textbf{MD2} & \textbf{CV} 
           & \textbf{LG} & \textbf{MD2} & \textbf{CV}
           & \textbf{LG} & \textbf{MD2} & \textbf{CV} \\
\midrule
P01  & 0.628 & 0.591 & 0.589 & 0.626 & 0.582 & 0.588 & 0.641 & 0.593 & 0.591 \\
P02  & 0.532 & 0.512 & 0.489 & 0.534 & 0.533 & 0.484 & 0.581 & 0.573 & 0.481 \\
P03  & 0.614 & 0.675 & 0.615 & 0.614 & 0.688 & 0.621 & 0.615 & 0.677 & 0.627 \\
P04  & 0.682 & 0.733 & 0.777 & 0.645 & 0.693 & 0.758 & 0.627 & 0.672 & 0.753 \\
P05  & 0.500 & 0.494 & 0.409 & 0.502 & 0.486 & 0.475 & 0.472 & 0.478 & 0.510 \\
P06  & 0.620 & 0.859 & 0.852 & 0.637 & 0.793 & 0.850 & 0.643 & 0.747 & 0.857 \\
P07  & 0.631 & 0.611 & 0.576 & 0.609 & 0.608 & 0.571 & 0.592 & 0.591 & 0.573 \\
P08  & 0.520 & 0.485 & 0.468 & 0.586 & 0.549 & 0.475 & 0.618 & 0.588 & 0.470 \\
P09  & 0.591 & 0.555 & 0.431 & 0.593 & 0.571 & 0.432 & 0.591 & 0.567 & 0.442 \\
P10  & 0.279 & 0.379 & 0.249 & 0.333 & 0.428 & 0.240 & 0.415 & 0.525 & 0.238 \\
P11  & 0.640 & 0.678 & 0.693 & 0.632 & 0.661 & 0.695 & 0.631 & 0.655 & 0.681 \\
P12  & 0.726 & 0.701 & 0.648 & 0.737 & 0.755 & 0.621 & 0.731 & 0.741 & 0.617 \\
\midrule
\textbf{Mean} & \textbf{0.580} & \textbf{0.606} & \textbf{0.566} 
              & \textbf{0.587} & \textbf{0.612} & \textbf{0.567}
              & \textbf{0.596} & \textbf{0.617} & \textbf{0.570} \\
\bottomrule
\end{tabular}
\end{table*}

\noindent\textbf{Q4 definition.}
For each predictor and participant, we compute the geodesic 
nonconformity score $s_i = d_{\text{geo}}(\hat{y}_i, y_i)$ 
(Eq.~\ref{eq:geo}) on the test set and partition frames into 
quartiles by score magnitude. Q4 denotes the hardest 25\% of 
frames meaning those with the largest geodesic nonconformity score. 
We validate this computationally over all test frames: for each 
quartile we compute mean Euclidean camera displacement 
$\|t_i - t_{i-1}\|$ from ground-truth EPIC-Fields trajectories, 
where $t_i$ is the ground-truth translation at frame $i$. Q4 
frames consistently exhibit higher displacement than Q1--Q3 
frames across participants and predictors (e.g., P07: Q4 mean 
1.69 vs.\ Q1--Q3 mean 0.70; P08: Q4 mean 3.70 vs.\ Q1--Q3 
mean 2.04), confirming they correspond to fast, difficult 
camera motions.

\subsection{Conditional Coverage Gap}
\label{sec:gap}

Table~\ref{tab:story1} reports within-participant Q4 coverage 
for camera pose prediction across all 12 participants and 3 
predictors at the $k^{th}$ future frame ($k \in \{10, 20, 30\}$).
To investigate whether the Q4 coverage gap is driven by 
participants with inherently fast motion or is a universal 
failure of single-threshold CP, we stratify participants by 
motion level based on their median const-vel calibration 
score: High (median $>$50: P01, P06--P08, P10, P12) and Low 
($<$50: P02--P05, P09, P11). These thresholds are in the 
units of Eq.~\ref{eq:geo} and correspond to a natural gap 
in the distribution of per-participant median scores 
(Low group: 8.0--34.5; High group: 70.9--141.1).
We define Standard CP as split CP with geodesic SE(3) 
nonconformity score and a single fixed threshold. This achieves 
nominal overall coverage ($\sim$0.91) but fails consistently 
on Q4, with coverage ranging from 0.25 to 0.85 and never 
reaching the 0.90 target. Crucially, the gap is not specific 
to high-motion participants: it is universal across all 108 participant-predictor-prediction 
horizon combinations, confirming it is 
a fundamental property of single-threshold CP rather than an 
artifact of any particular predictor or motion regime.

\subsection{Geodesic vs.\ Euclidean Scoring}
\label{sec:geo_vs_euc}

We compare geodesic and Euclidean nonconformity scores across 
all 10 participants with available residuals, reporting detailed 
coverage for a representative cross-participant split (cal=P06, 
test=P07+P08), chosen as the three highest-median-residual 
participants ($\tilde{s}=141.1$, 86.2, 112.7 respectively).

The two scores identify systematically different hard frames:
geodesic and Euclidean Q4 sets disagree 74--75\% of the time 
across participants for both LightGlue and MonoDepth2. To 
determine which set corresponds to genuinely hard frames, we 
measure mean ground-truth displacement $\|t_i - t_{i-1}\|$. 
Geodesic Q4 frames have 2--3$\times$ larger displacement than 
Euclidean Q4 frames (P07: 1.69 vs.\ 0.57, P08: 3.70 vs.\ 2.32), 
confirming that geodesic scoring identifies physically hard frames 
while Euclidean scoring does not. Consequently, geodesic CP 
achieves Q4 coverage of 0.873 vs.\ Euclidean CP's 0.704 for 
LightGlue at $k=10$; results for $k=20,30$ follow the same 
trend (Table~\ref{tab:story2}).
\begin{table}[h]
\centering
\caption{Cross-participant Q4 coverage: geodesic vs.\ Euclidean 
(cal=P06, test=P07+P08). Q4 defined by each score's own residuals.}
\label{tab:story2}
\setlength{\tabcolsep}{4pt}
\begin{tabular}{lcccccc}
\toprule
& \multicolumn{2}{c}{$k=10$} & \multicolumn{2}{c}{$k=20$} & 
\multicolumn{2}{c}{$k=30$} \\
\cmidrule(lr){2-3}\cmidrule(lr){4-5}\cmidrule(lr){6-7}
\textbf{Predictor} & \textbf{Geo} & \textbf{Euc} & \textbf{Geo} 
& \textbf{Euc} & \textbf{Geo} & \textbf{Euc} \\
\midrule
LightGlue   & 0.873 & 0.705 & 0.833 & 0.700 & 0.807 & 0.701 \\
MonoDepth2  & 0.605 & 0.524 & 0.567 & 0.550 & 0.565 & 0.553 \\
Const-vel   & 0.647 & 0.647 & 0.639 & 0.638 & 0.633 & 0.632 \\
\bottomrule
\end{tabular}
\end{table}

\subsection{Adaptive CP via DINOv2-Bridge}
\label{sec:adaptive}

We replace the single fixed threshold of standard CP with a 
per-frame difficulty-weighted threshold, adapting~\citet{shahbazi2026} 
from $\mathbb{S}^2$ to SE(3) egocentric pose with the additional 
challenge of cross-participant generalization: the difficulty 
estimator must transfer to unseen participants without any RGB 
frames at deployment time.

We estimate per-frame difficulty $\sigma$ via a two-stage pipeline.
Kinematic features alone are an incomplete difficulty signal: a 
frame can have slow motion yet be visually unlocalizeable due to 
reflective surfaces or texture-poor regions. \textbf{Stage~1 
(DINOv2-MLP)} maps RGB frames to predicted geodesic nonconformity 
score magnitude $\hat{s}$, trained on P01 (56K frames) at $k=10$; 
the learned signal generalizes across horizons since visual 
difficulty is a property of the frame, not the prediction horizon. 
\textbf{Stage~2 (Bridge MLP)} distills this into 20-dim kinematic 
pose features (velocity, acceleration, higher-order motion 
statistics), enabling image-free difficulty estimation on unseen 
participants at test time.

We evaluate on the constant-velocity predictor, as kinematic 
residuals align naturally with the Bridge's input features; 
extension to LightGlue and MonoDepth2 is left to future work.
Table~\ref{tab:story3} shows Bridge closes Q4 from 0.73--0.78 
to 0.88--0.93 across both splits and all $k$, while maintaining 
overall coverage at the 90\% target.
\begin{table}[h]
\centering
\caption{Adaptive CP results (const-vel, $\alpha=0.10$, Train=P01). 
Overall (Ovr) and Q4 coverage. Bridge (ours).}
\label{tab:story3}
\setlength{\tabcolsep}{4pt}
\small
\begin{tabular}{lcccccc}
\toprule
& \multicolumn{2}{c}{$k=10$} & \multicolumn{2}{c}{$k=20$} 
& \multicolumn{2}{c}{$k=30$} \\
\cmidrule(lr){2-3}\cmidrule(lr){4-5}\cmidrule(lr){6-7}
\textbf{Method} & \textbf{Ovr} & \textbf{Q4} 
& \textbf{Ovr} & \textbf{Q4} 
& \textbf{Ovr} & \textbf{Q4} \\
\midrule
\multicolumn{7}{l}{\textit{Cal=P06, Test=P07, P08, P10, P12}} \\
\midrule
Std CP        & 0.938 & 0.751 & 0.935 & 0.738 & 0.932 & 0.728 \\
Bridge (ours) & 0.935 & \textbf{0.929} & 0.931 & \textbf{0.924} 
              & 0.927 & \textbf{0.893} \\
\midrule
\multicolumn{7}{l}{\textit{Cal=P10, Test=P06, P07, P08, P12}} \\
\midrule
Std CP        & 0.939 & 0.754 & 0.942 & 0.768 & 0.945 & 0.780 \\
Bridge (ours) & 0.929 & \textbf{0.934} & 0.934 & \textbf{0.928} 
              & 0.945 & \textbf{0.879} \\
\bottomrule
\end{tabular}
\end{table}

\section{Discussion}
\label{sec:discussion}

\noindent\textbf{Limitations \& Future Work.}
DINOv2-Bridge is evaluated only on the constant-velocity predictor;
extending to LightGlue and MonoDepth2 is non-trivial due to 
residual scale mismatch and remains future work. Bridge also relies 
on P01 as sole training source, and performance may vary with 
different participants or environments. Natural extensions include 
applying Bridge to stronger predictors 
(DROID-SLAM~\cite{teed2021}, MAC-VO~\cite{macvo2024}) and 
evaluating on Ego-Exo4D~\cite{grauman2024} for cross-domain 
generalization.

%% file: main.bib
@String(CVPR= {IEEE Conf. Comput. Vis. Pattern Recog.})

@String(ICCV= {Int. Conf. Comput. Vis.})

@String(CVPR  = {CVPR})

@String(ICCV  = {ICCV})

@book{vovk2005,
  author    = {Vovk, Vladimir and Gammerman, Alex and Shafer, Glenn},
  title     = {Algorithmic Learning in a Random World},
  publisher = {Springer},
  year      = {2005}
}

@article{angelopoulos2022,
  author  = {Angelopoulos, Anastasios N. and Bates, Stephen},
  title   = {A Gentle Introduction to Conformal Prediction and 
             Distribution-Free Uncertainty Quantification},
  journal = {Foundations and Trends in Machine Learning},
  volume  = {16},
  number  = {4},
  pages   = {494--591},
  year    = {2023}
}

@inproceedings{yang2023,
  author    = {Yang, Heng and Pavone, Marco},
  title     = {Object Pose Estimation with Statistical Guarantees: 
               Conformal Keypoint Detection and Geometric 
               Uncertainty Propagation},
  booktitle = {CVPR},
  pages     = {8947--8958},
  year      = {2023}
}

@inproceedings{closure2024,
  author    = {Yang, Heng and Pavone, Marco},
  title     = {{CLOSURE}: Fast Quantification of Pose Uncertainty Sets},
  booktitle = {Robotics: Science and Systems},
  year      = {2024}
}

@inproceedings{stutts2023,
  author    = {Stutts, Alex C. and Erricolo, Danilo and 
               Tulabandhula, Theja and Trivedi, Amit Ranjan},
  title     = {Lightweight, Uncertainty-Aware Conformalized 
               Visual Odometry},
  booktitle = {CVPR Workshops},
  year      = {2023}
}

@article{shahbazi2026,
  author  = {Amiri Shahbazi, Marzieh and Baheri, Ali},
  title   = {Geometry-Aware Uncertainty Quantification via 
             Conformal Prediction on Manifolds},
  journal = {arXiv:2602.16015},
  year    = {2026}
}

@inproceedings{epicfields2023,
  author    = {Tschernezki, Vadim and Sherburn, Ahmad and 
               Davison, Andrew J. and Damen, Dima},
  title     = {{EPIC-Fields}: Marrying {3D} Geometry and 
               Video Understanding},
  booktitle = {NeurIPS},
  year      = {2023}
}

@article{damen2022,
  author  = {Damen, Dima and Doughty, Hazel and Farinella, 
             Giovanni Maria and others},
  title   = {Rescaling Egocentric Vision: Collection, Pipeline 
             and Challenges for {EPIC-KITCHENS-100}},
  journal = {International Journal of Computer Vision},
  volume  = {130},
  pages   = {33--55},
  year    = {2022}
}

@inproceedings{lindenberger2023,
  author    = {Lindenberger, Philipp and Sarlin, Paul-Edouard and 
               Pollefeys, Marc},
  title     = {{LightGlue}: Local Feature Matching at Light Speed},
  booktitle = {ICCV},
  year      = {2023}
}

@inproceedings{godard2019,
  author    = {Godard, Cl{\'e}ment and Mac Aodha, Oisin and 
               Firman, Michael and Brostow, Gabriel},
  title     = {Digging into Self-Supervised Monocular Depth 
               Estimation},
  booktitle = {ICCV},
  year      = {2019}
}

@inproceedings{romano2019,
  author    = {Romano, Yaniv and Patterson, Evan and 
               Cand{\`e}s, Emmanuel},
  title     = {Conformalized Quantile Regression},
  booktitle = {NeurIPS},
  year      = {2019}
}

@inproceedings{teed2021,
  author    = {Teed, Zachary and Deng, Jia},
  title     = {{DROID-SLAM}: Deep Visual {SLAM} for Monocular, 
               Stereo, and {RGB-D} Cameras},
  booktitle = {NeurIPS},
  year      = {2021}
}

@inproceedings{grauman2024,
  author    = {Grauman, Kristen and Westbury, Andrew and others},
  title     = {Ego-{Exo4D}: Understanding Skilled Human Activity 
               from First- and Third-Person Perspectives},
  booktitle = {CVPR},
  year      = {2024}
}

@inproceedings{macvo2024,
  author    = {Wang, Yuheng and others},
  title     = {{MAC-VO}: Metrics-Aware Covariance for 
               Learning-Based Stereo Visual Odometry},
  booktitle = {ICRA},
  year      = {2025}
}

@misc{pan2024globalstructurefrommotionrevisited,
      title={Global Structure-from-Motion Revisited}, 
      author={Linfei Pan and Dániel Baráth and Marc Pollefeys and Johannes L. Schönberger},
      year={2024},
      eprint={2407.20219},
      archivePrefix={arXiv},
      primaryClass={cs.CV},
      url={https://arxiv.org/abs/2407.20219}, 
}
